\pdfoutput=1

\documentclass[11pt]{article}

\usepackage[]{acl}

\usepackage{times}
\usepackage{latexsym}

\usepackage{microtype}

\usepackage[inline]{enumitem}
\usepackage[utf8]{inputenc} 
\usepackage[T1]{fontenc}    
\usepackage[russian,english]{babel}
\usepackage{hyperref}       
\usepackage{url}            
\usepackage{amsfonts}       
\usepackage{nicefrac}       
\usepackage{microtype}      
\usepackage{xcolor}         

\usepackage{booktabs}
\usepackage{amsmath}
\usepackage{multirow}
\usepackage{graphicx}
\usepackage{enumitem}
\usepackage{subcaption}
\usepackage{caption}
\usepackage{rotating}
\usepackage[normalem]{ulem}
\usepackage{amssymb}
\usepackage{colortbl}
\usepackage{linguex}

\usepackage{siunitx} 
\usepackage{multirow, makecell}
\usepackage{pifont}
\usepackage{siunitx} 
\usepackage{multirow, makecell}
\usepackage{tabularx}
\usepackage[all]{nowidow}
\definecolor{darkgreen}{rgb}{0.0, 0.42, 0.24}
\definecolor{green}{RGB}{112, 173,71}
\definecolor{blue}{RGB}{68, 114,196}
\definecolor{orange}{RGB}{237, 125,49}
\definecolor{red}{RGB}{202, 54,49}
\definecolor{yellow}{RGB}{222,194, 142}

\newcommand{\mt}{\textsc{MT}\xspace}
\newcommand{\wmt}{\textsc{WMT}\xspace}
\newcommand{\xglm}{\textsc{XGLM}\xspace}
\newcommand{\nlp}{\textsc{NLP}\xspace}
\newcommand{\plm}{\textsc{PLM}\xspace}
\newcommand{\incontext}{In-context\xspace}

\newcommand{\knnmt}{\textit{k}\textsc{NN-MT}\xspace}

\newcommand{\ngram}{n-gram\xspace}

\newcommand{\bm}{\textsc{BM25}\xspace}
\newcommand{\bleu}{\textsc{Bleu}\xspace}

\usepackage[linesnumbered,ruled,vlined]{algorithm2e}
\SetKwInput{KwInput}{Input} 
\SetKwInOut{KwOutput}{Output} 

\definecolor{hotpink}{HTML}{EF7C8E}
\definecolor{tiffanyblue}{HTML}{A0E7E5}
\definecolor{mint}{HTML}{B4F8C8}
\definecolor{paleyellow}{HTML}{FBE7C6}
\definecolor{rosewater}{HTML}{D8A7B1}
\definecolor{cream}{HTML}{FAE8E0}

\DeclareMathOperator*{\argmax}{arg\,max}

\title{\incontext Examples Selection for Machine Translation}

\author{Sweta Agrawal$^1$\thanks{\quad Work done during internship at Meta AI Research.}, Chunting Zhou$^2$, Mike Lewis$^2$, Luke Zettlemoyer$^2$, Marjan Ghazvininejad$^2$  \\
$^1$ University of Maryland \quad  \quad $^2$ Meta AI \\
 \texttt{sweagraw@umd.edu}  \quad \quad \texttt{\{chuntinz,mikelewis,lsz,ghazvini\}@meta.com}
}

\begin{document}
\maketitle
\begin{abstract}

Large-scale generative models show an impressive ability to perform a wide range of Natural Language Processing (\nlp) tasks using \MakeLowercase{\incontext} learning, where a few examples are used to describe a task to the model. For Machine Translation (\mt), these examples are typically randomly sampled from the development dataset with a similar distribution as the evaluation set. However, it is unclear how the choice of these \MakeLowercase{\incontext} examples and their ordering impacts the output translation quality. In this work, we aim to understand the properties of good \MakeLowercase{\incontext} examples for \mt in both in-domain and out-of-domain settings. We show that the translation quality and the domain of the \MakeLowercase{\incontext} examples matter and that 1-shot noisy unrelated example can have a catastrophic impact on output quality. While concatenating multiple  random examples reduces the effect of noise, a single \textit{good} prompt optimized to maximize translation quality on the development dataset can elicit learned information from the pre-trained language model. Adding similar examples based on an \ngram overlap with the test source significantly and consistently improves the translation quality of the outputs, outperforming a strong \knnmt baseline in 2 out of 4 out-of-domain datasets.  


\end{abstract}

\section{Introduction}

\incontext learning \cite{brown2020language} has recently received a lot of attention from the \nlp research community due to its remarkable ability to utilize only a few input-output examples to perform many \nlp tasks \cite{promptsurvey2021liu}. For example, \citet{lin2021few} demonstrate that a 7.5B multilingual generative model, \xglm, outperforms a supervised sequence-to-sequence baseline in 45 translation directions on the FLORES-101 machine translation benchmark \cite{goyal2022flores} using just $32$ randomly sampled translation examples as demonstrations. While these results are compelling, recent work has also shown that the performance and capability of a pre-trained language model (\plm) can be highly sensitive to many factors, such as the choice of \MakeLowercase{\incontext} examples \cite{liu-etal-2022-makes}, their ordering \cite{lu2022fantastically} and the template \cite{jiang-etal-2020-know}.

Typically, \MakeLowercase{\incontext} learning for \mt uses examples that are randomly sampled from a small development set that resembles the domain of the test dataset. The effect of the aforementioned factors (such as the choice of the examples) on the translation quality of the \plm hence remains unclear and unexplored. Yet another crucial gap in using \MakeLowercase{\incontext} learning for \mt in the current literature is the effect of the domain of \MakeLowercase{\incontext} examples on translation quality since out-of-domain generalization is a known and important challenge in \mt \cite{koehn2017six}. 

In this work, we systematically analyze how factors such as the choice and the number of few-shot \MakeLowercase{\incontext} examples and their ordering impact \mt output quality. We show that while noisy unrelated 1-shot example can have a significantly adverse effect on translation quality, a single prompt optimized to maximize the translation quality on a development set can sufficiently elicit task-based information from the \plm. Our analysis thus demonstrates the importance of selecting good examples for \mt and raises the question: \textit{What are the properties of good \MakeLowercase{\incontext} examples for \mt?} In that direction, our findings suggest that a well-formed meaning-equivalent translation example results in higher quality translation than randomly selected \MakeLowercase{\incontext} examples. 

Furthermore, motivated by the use of Translation Memory in Computer-Aided Translation \cite{yamada2011effect} and its usage in computational approaches to Machine Translation \citep[\textit{inter alia}]{somers1999example, koehn2010convergence, khandelwal2020nearest}, we retrieve similar examples to the test source from a datastore that includes pairs of source text
and their corresponding translations via \bm, an unsupervised efficient retriever to provide additional context to the model. As the context window of the \plm is usually limited ($\sim$ 3096 tokens, $16-20$ examples), we propose a novel \MakeLowercase{\incontext}-example selection and reranking strategy that maximizes the coverage of the source n-grams in the selected examples. 
Experiments on \textsc{WMT'19} English$\leftrightarrow$German and English$\leftrightarrow$Russian datasets show that our proposed re-ranking strategy can consistently improve the translation quality over the outputs generated using \bm retrieved examples. 

Combining optimized 1-shot task-level with example-specific \MakeLowercase{\incontext} examples using a simple concatenation strategy further improves translation quality, outperforming state-of-the-art inference-adapted nearest-neighbor \mt models (\knnmt) on two out-of-domain datasets (Medical and IT) while being memory and compute efficient as our approach does not require constructing and querying a dense token-level datastore.


\section{Background: \incontext Learning} 
Generating translations from large-scale multilingual language models like m\textsc{GPT} \cite{shliazhko2022mgpt}, \textsc{XGLM} \cite{lin2021few} or AlexaTM 20B \cite{soltan2022alexatm} requires conditioning the decoder-only language model with in-context parallel examples. These examples serve two purposes: a) providing the model with the format and knowledge of the task (\textbf{task-level}) and b) guiding the output generation via providing useful information about the unseen source sentence (\textbf{example-specific}). 
This is different from the standard sequence-to-sequence models, where the task is always known, and the model learns generalizable patterns from the input-output examples to perform the task (in this case, translation) for the unseen source text.

\begin{table}[h]
    \centering
        \renewcommand\tabularxcolumn[1]{m{#1}}
        \renewcommand\arraystretch{1.2}
    \begin{tabularx}{\columnwidth}{*{1}{>{\arraybackslash}X}}
  \underline{Source:} \textit{Welche Risiken sind mit} \textbf{Poulvac FluFend H5N3 RG} \textit{verbunden?} \\ 
   \addlinespace[0.1cm]
    \underline{Template:} \{Source text\} = \{Target text\}. \\
     \addlinespace[0.1cm]
   \underline{Example-Specific:} \textit{Welche Risiken sind mit} Sebivo \textit{verbunden?} = What are the risks associated with Sebivo? \\
   \underline{Task-Level:} Bei PROMESS1 werden drei Hauptziele verfolgt. = PROMESS1 has three main objectives. \\
    \end{tabularx}  
    \caption{\incontext Examples for Machine Translation.} \label{tab:example}
\end{table}

Formally, given $k$ in-context examples $\{x_i, y_i\}_{1}^{k}$ the prefix input or the prompt, $x_j^p$, is generated by concatenating the demonstration examples $\{(x_i, y_i)\}_{1}^{k}$ to the test input, $x_j^s$ according to a \textit{template}, $P$ (see Table~\ref{tab:example}). The output, $\hat{y}$, is then generated via the \plm with parameters $\theta$ via greedy decoding as follows:
\begin{equation} \label{eq:generate}
    \hat{y}_{j,t} = \argmax_{y'_{j,t}} P_{\plm}(y'_{j,t} | x_j^p, \hat{y}_{j,<t}; \theta)
\end{equation}

\section{Prompt Selection}

Ideally, good \MakeLowercase{\incontext} examples can trigger the pre-trained language model to generate the \textbf{desired output} and also elicit the information learned during pre-training \cite{jiang-etal-2020-know}. \citet{min2022rethinking} show that, for classification tasks, the \MakeLowercase{\incontext} examples provide information about the task (the distribution of the input text, the label space, and the format of the task) and that the model does not rely on these examples to generate the final output. However, their analysis is limited to a) classification tasks and 2) randomly sampled \MakeLowercase{\incontext} examples. Prior work has also shown that the order of these \MakeLowercase{\incontext} examples can also lead to high variance in downstream performance \cite{zhang-etal-2022-prompt}. However, less is understood about how these factors impact text generation tasks like \mt. \textit{Do we need multiple \MakeLowercase{\incontext} examples?} \textit{What makes good \MakeLowercase{\incontext} examples for \mt?} 
\textit{How sensitive is the model to the order of the prompts?} 

In this work, we aim to better understand the impact of prompt selection on the translation quality of the outputs. Given a training dataset consisting of $n$ parallel examples $D=\{x_i, y_i\}_{i=1}^{n}$, and a test source $x_j$, we select a subset of $m$ \textit{informative} samples to form a prompt which either provides task-level and/or example-specific information as discussed below.

\begin{figure*}[ht!]
  \centering
  \includegraphics[width=\linewidth]{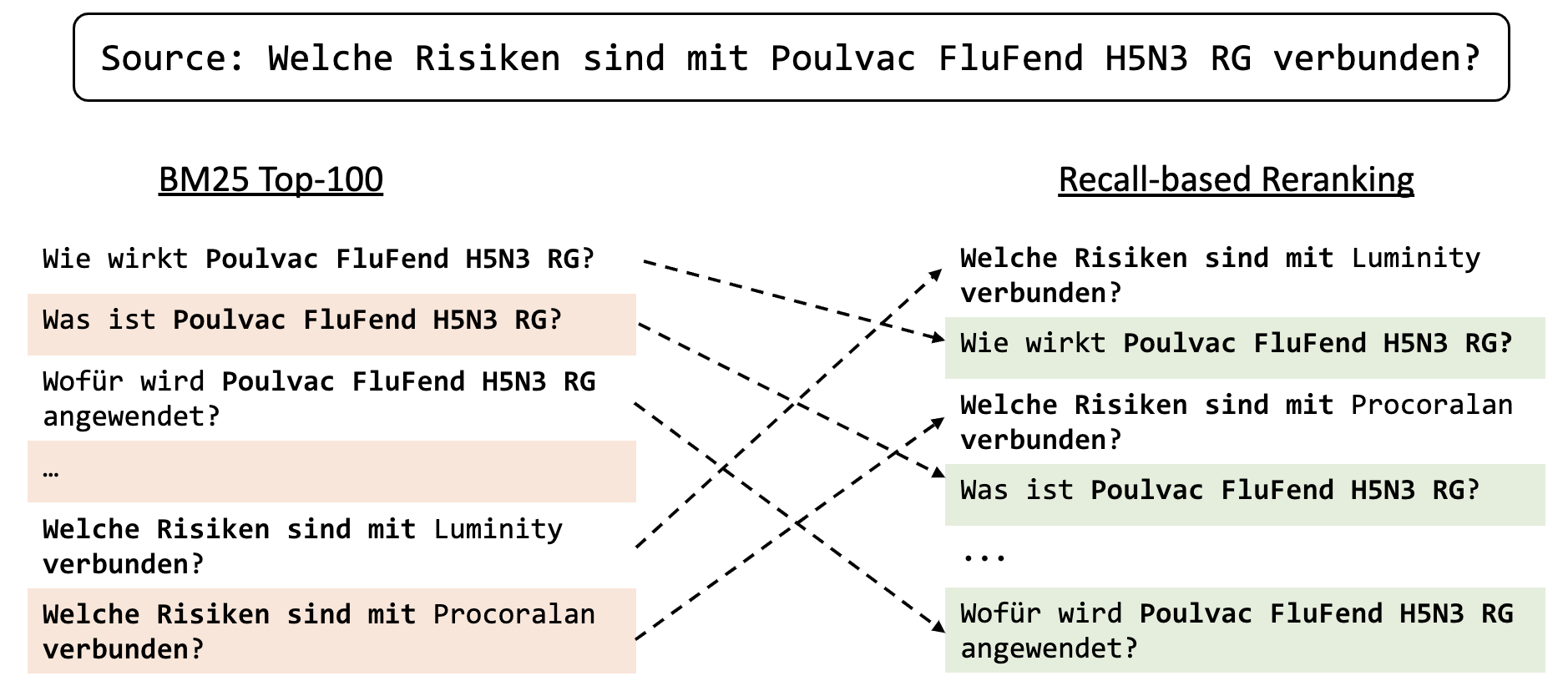}
\caption{Our proposed strategy can cover all the terms from the input text,``Welche Risiken sind mit Poulvac FluFend H5N3 RG verbunden?'', in this case with just the two examples. }  \label{fig:rerank}
\end{figure*}

\begin{algorithm*}
\KwInput{Prompts $\{P_j(x_i, y_i)\}_{1}^{k}$ for the test source $x_j^s$, $\lambda$, Threshold}
\KwOutput{Ordered Selected Prompts $\{T = P_j(x_i, y_i)\}_{1}^{s}$, s $\leq$ k}
$T\leftarrow$ Empty Ordered List \\
$ S  \leftarrow \colorbox{rosewater}{EXTRACT\_WORD\_NGRAMS\_WITH\_COUNTS} (x_j^s)$ \label{ref:eqsrcngram} \\ 
\For{$i \in \{1..k\}$}
{
    $Q[i]  \leftarrow \colorbox{rosewater}{EXTRACT\_WORD\_NGRAMS\_WITH\_COUNTS} (P_j (x_i))$ \label{ref:eqhypngram} \\ 
}
\While{True}
{
    \For{$i \in \{1..k\}$}
    {
        $Score[i] \leftarrow \colorbox{cream}{NGRAM\_OVERLAP\_SCORE} (S, Q[i])$ \label{ref:eqngramscore}
    }
    \If{$max(Score) < Threshold$}{$break$} 
    $T.append(P_{argmax(Score)})$ \\
    $matched\_ngrams \leftarrow S \cap Q[argmax(Score)]$ \\
    $Q[argmax(Score)] \leftarrow \emptyset$  \\
    \For{$ngram \in matched\_ngrams$}
    {
        $Count_{S}(ngram) \times= \lambda$ \label{ref:eqdownweighengram}
    }
}
Return $T$
\caption{An N-gram Recall-based Strategy to Re-rank \incontext Examples} \label{algorithm}
\end{algorithm*}  

\subsection{Task-level \incontext Examples} \label{subsec:task}

A good task-level \MakeLowercase{\incontext} example should be able to elicit information learned during pretraining from the \plm. One way to measure the efficacy of an example as a prompt is via computing the translation quality of the outputs generated when prompting the \plm given an example. Hence, we select the task-level prompt as follow: For a given example sampled from the training dataset, $(x_i, y_i) \in D^S$, we create a prompt, 
$x_i^p$ by concatenating the example $\{(x_i, y_i)\}$ to each source in the development set. The system outputs are then generated using equation~\ref{eq:generate}. We then rank examples from $D^S$ as task-level prompts based on the \bleu of the generated outputs against the references on this held-out development set, $D^{dev}=\{X, Y\}$: 

\begin{equation}
    (x_s, y_s) = \argmax_{(x, y) \in D^S} \bleu (Y, \hat{Y})
\end{equation}

\subsection{Example-specific \incontext Examples} \label{subsec:example}

Prior work on retrieving \textit{good} \MakeLowercase{\incontext} example-specific prompts for tasks other than \mt (like question answering or knowledge retrieval) either trains a dense-retriever \cite{rubin2021learning} or utilizes samples that are closer to the \textbf{test source} in the embedding space of a \plm like \textsc{BERT} \cite{devlin-etal-2019-bert}, Ro\textsc{BERT}a \cite{liu2019roberta}, or \textsc{XLNet} models \cite{liu-etal-2022-makes}. While contextual models can generate a global sentence representation, they overlook rare lexicons which can be important for generating translations in unseen domains like medical or IT \cite{wrzalik-krechel-2021-cort}. 

However, for \mt, overlapping n-grams between the source and the retrieved sentences ensures informativeness as the target associated with the retrieved sentence is likely to include partial translations of the source. We can thus use \bm as an efficient unsupervised retrieval method to retrieve similar examples. However, as the examples are scored independently and \bm favors rare word matches \cite{robertson2009probabilistic}, the top retrieved candidates might not cover all the terms in the source text (Figure~\ref{fig:rerank}). Given that the context window of the \plm is usually limited ($\sim$ 3096 tokens, $16-20$ examples), maximizing the coverage of all the terms found in the test input might be favorable.  Hence, we propose to re-rank the top $100$ candidates retrieved from \bm using our algorithm outlined in \ref{algorithm}. 

We extract all the word n-grams, and their counts from the test source, $x_j^s$ and source of the \bm retrieved examples, $\{P_j(x_i\}_{1}^{k}$ (lines ~\ref{ref:eqsrcngram}-\ref{ref:eqhypngram}). Let S and Q denote the set of the source n-grams and the n-grams from a \bm retrieved example, respectively. We compute a recall-based (R) n-gram overlap score (line~\ref{ref:eqngramscore}) using the following equation:

\begin{align}
    R_{n} &= \frac{\sum_{ngram \in S \cap Q} Count_{matched} (ngram)}{\sum_{ngram \in S} Count_{S}(ngram)} \\
    Score &= \exp(\frac{1}{4} \sum_n \log(R_{n}))
\end{align}

The example with the maximum score is then added to the set of selected prompts, and the found n-grams from the test source are then down-weighted by a factor, $\lambda$ for the next iteration of selection (line~\ref{ref:eqdownweighengram}). For example, setting $\lambda=0$ will select the example that covers the n-grams from the test source in the subsequent iteration that has not already been encountered. This process is then repeated over the retrieved pool until a set threshold of the score is reached. 

Figure~\ref{fig:rerank} shows the top-100 candidates retrieved via \bm for the input: ``Welche Risiken sind mit Poulvac FluFend H5N3 RG verbunden?''. The top few candidates provide the same information to the \plm, i.e., translation of the phrase ``Poulvac FluFend H5N3 RG''. The examples including the other terms (``Welche Risiken sind mit verbunden ?'') from the input text, are ranked lower. On the other hand, our proposed re-ranking strategy can cover all the terms from the input text, in this case, with just the top-2 examples.

%


\section{Evaluation Settings}
\subsection{Datasets and Evaluation Metric} We perform our in-domain evaluation on the WMT-19 German (de) $\Leftrightarrow$ English (en) and WMT-19 Russian (ru) $\Leftrightarrow$ English (en) datasets \cite{barrault-etal-2019-findings}. For the out-of-domain evaluation, we use the multi-domain dataset from \citet{aharoni-goldberg-2020-unsupervised} for the following domains: Medical, Law, IT, and Koran. The dataset statistics are included in the Appendix (Table~\ref{tab:ablation_datastats}). We normalize punctuation using the Moses toolkit \cite{koehn-etal-2007-moses} and remove sentences longer than 250 tokens as well as sentence pairs with a source/target length ratio exceeding 1.5 from the in-domain datasets. We evaluate the detokenized length truncated outputs generated by the model using sacreBLEU \cite{post-2018-call}.\footnote{\url{https://github.com/mjpost/sacrebleu} \\ We also report Comet \cite{rei-etal-2020-comet} scores for evaluating translation quality in Appendix Tables~\ref{tab:comet_ind_eval} and ~\ref{tab:comet_ood_eval}.} The generated outputs from the \plm are truncated to twice the source length, as preliminary analysis suggested degeneration in a few ($\sim$10-20) examples.

\begin{table*}[t]
\centering
\scalebox{0.90}{
\begin{tabular}{lcrrrrr}
 \toprule
\bf{Method}& $\bf{p+q_{max}}$ &  {\bf{En-De}}  & {\bf{De-En}}& {\bf{Ru-En}}   &  {\bf{En-Ru}} & \bf{Avg.} \\ 
  \midrule
Task-level & \colorbox{pink}{$1+0$}  & 23.35 & 32.16  & 30.48  & 25.04 & 27.75 \\
 \addlinespace[0.1cm]
 \bm& \colorbox{pink}{$0+1$} & 19.17  & 25.82& 24.54 &21.51 & 22.76\\
 \addlinespace[0.1cm]
R-\bm & \colorbox{pink}{$0+1$}     & 20.60& 28.19 & 27.26 &  21.92 & 24.49 \\

\addlinespace[0.1cm]
\midrule
\addlinespace[0.1cm]

 
   
Random (\textit{Baseline}) & \colorbox{paleyellow}{$16+0$}    & 24.48& 31.26  & 30.38  & 25.67 & 27.95 \\
      \addlinespace[0.1cm]
Task-level  & \colorbox{paleyellow}{$16+0$}  &23.72&   31.22  & 30.89  & 27.27 & 28.28 \\
        \addlinespace[0.1cm]
 \bm   & \colorbox{paleyellow}{$0+16$} &26.58  & 32.16&  31.44 &28.54 & 29.68\\
    \addlinespace[0.1cm]
  R-\bm  & \colorbox{paleyellow}{$0+16$} & 27.07 & 32.59 & 31.85 &28.90 & 30.10\\
  
\addlinespace[0.1cm]
\midrule
\addlinespace[0.1cm]

   R-\bm  & \colorbox{mint}{$0+17$} & 27.00 & 32.68 & 31.88 &28.80 & 30.09\\
   \addlinespace[0.1cm]
Task-level + R-\bm &   \colorbox{mint}{$1+16$}  & \bf{27.09} & \bf{33.24}& \bf{31.90} &  \bf{29.50} & \bf{30.43} \\
 
  \bottomrule
 \end{tabular}
 }
\caption{Results on \textsc{WMT}'19 test sets: Concatenating task-level prompt to R-\bm consistently achieves the best \bleu scores across the board. p and $q_{max}$ are the number of task-level and example-specific prompts respectively.}\label{tab:main_ind_eval} 

\end{table*}

\subsection{Experimental Conditions}

\paragraph{Language Model} We use the publicly available checkpoint of the $\text{XGLM}_{7.5B}$, a decoder-only causal language multilingual model \cite{lin2021few} for all our experiments, which has 32 layers and a hidden dimension of 4096. 

\paragraph{Baselines and Comparisons}
We consider the following comparisons:

\begin{itemize}[leftmargin=*]
    \item \textbf{Random}: $p$ random few-shot examples sampled from the training dataset (number of trials=3). 
    \item \textbf{Task-level}: top-$p$ examples that achieve the highest \bleu on the development set (\S~\ref{subsec:task}). 
    \item \textbf{Retrieved \incontext (\bm)}: $q_{max}$ examples retrieved via \bm, since unlike task-level examples, there is no guarantee that exactly $q$ similar examples will be found in the training dataset for each input.
    \item \textbf{Retrieved Re-ranked \incontext (R-\bm)}: $q_{max}$ re-ranked examples using our proposed approach as detailed in \S~\ref{subsec:example}. 
\end{itemize}

We additionally compare our results with \knnmt \cite{khandelwal2020nearest} for out-of-domain evaluation. We use $\lambda=0.1$, threshold=$1.0$ and order the examples according to their similarity to the source, with the most similar examples on the left in all our experiments based on an initial hyperparameter search on the development dataset (Appendix Tables~\ref{tab:order},\ref{tab:hyp_lambda}). 

%


\begin{table*}[t]
\centering
\scalebox{0.90}{
\begin{tabular}{llcrrrrr}
 \toprule
\bf{Method}& \bf{Corpus} & $\bf{p+q_{max}}$  & {\bf{MEDICAL}} & {\bf{LAW}} & {\bf{IT}}   &  {\bf{KORAN}}  &  {\bf{Avg.}}  \\ 
  \midrule
\multirow{2}{*}{Task-level} &  Domain-specific & \multirow{2}{*}{\colorbox{pink}{$1+0$}} & 31.23 & 32.10 & 28.70 & 14.68 & 26.68 \\
& \textsc{WMT}  &  &  30.08 & 31.10 &  26.72 & 13.19 & 25.27 \\
R-\bm &  Domain-specific & \colorbox{pink}{$0+1$} & 52.62 & 55.46  & 40.54 & 13.76 & 40.60\\

\addlinespace[0.1cm]
\midrule
\addlinespace[0.1cm]

\multirow{2}{*}{Task-level} & Domain-specific  & \multirow{2}{*}{\colorbox{paleyellow}{$16+0$}} & 32.65 & 33.68 & 28.81 & 15.30 & 27.61 \\
&  \textsc{WMT} &  & 30.14 & 30.76 & 26.19 & 12.72 & 24.95\\
   \addlinespace[0.1cm]
 R-\bm& Domain-specific  & \colorbox{paleyellow}{$0+16$} & 56.43 & \bf{59.57} & 46.57  & 17.49 & 45.02 \\

\addlinespace[0.1cm]
\midrule
\addlinespace[0.1cm]
     R-\bm   & \multirow{2}{*}{Domain-specific} & \colorbox{mint}{$0+17$} &  56.65 & 59.55 & 46.64 & 17.48 & 45.08 \\
     \addlinespace[0.1cm]
 Task-level + R-\bm &      &    \colorbox{mint}{$1+16$}  & \bf{56.76} & 59.56 & \bf{47.50} & \bf{17.55} & \bf{45.34} \\ 
  \addlinespace[0.1cm]
  \midrule
  \knnmt & - & -  & 54.35 & 61.78 & 45.82 & 19.45 & 45.35\\
  \bottomrule
 \end{tabular}
 }
\caption{Results on the Multi-Domain Test Set: Prompting \xglm with R-\bm \MakeLowercase{\incontext} examples outperforms \knnmt on 2 out of 4 domains.}\label{tab:main_ood_eval} 
\end{table*}


\section{Results}

Table~\ref{tab:main_ind_eval} and~\ref{tab:main_ood_eval} summarizes our main results for the in-domain evaluation when translating between English <-> German and English <-> Russian and the four German-English out-of-domain datasets, respectively.

\subsection{In-domain Evaluation}
\paragraph{A single task-level prompt is competitive with 16 random few-shot examples.} Our experiment suggests that it is possible to elicit the task-level knowledge from the large-scale language model using a single prompt as opposed to using 16 random few-shot examples when translating into English (Table~\ref{tab:main_ind_eval}). Using a single task-level prompt (optimized on the development set) improves \bleu over using $16$ random few-shot examples for 2 out of 4 translation directions (De-En, Ru-En). We hypothesize that when translating out of English, the model still benefits from getting exposed to multiple and diverse random few-shot examples as the target language model is relatively weaker.

\paragraph{Multiple example-specific prompts are required to improve translation quality over a single task-level prompt. } Using a single task-level ($p=1$) prompt attains higher \bleu over using a single example-specific prompt ($q=1$; \bm, R-\bm) across the board. By contrast, using upto $16$ \bm prompts ($q_{max}=16$) significantly improves output quality over using task-level prompts, with an average gain of $1.41$ in \bleu. 

\paragraph{Re-ranking \bm retreived examples improves \bleu.} Our proposed re-ranking strategy consistently improves \bleu across the board over \bm for both values of $q_{max}=\{1, 16\}$ showing that both the order and the choice of the \MakeLowercase{\incontext} examples matters. 

Both task-level and R-\bm examples provide complementary advantages, as combining them using a simple concatenation strategy improve output quality over task-level or R-\bm examples. We leave the exploration of optimizing the number and the joint order of task-level and example-specific prompts to future work.


\subsection{Out-of-domain Evaluation}

As \xglm is trained on monolingual Common Crawl snapshots, translation in any domain and language could be considered an out-of-domain task from the model's perspective. However, we hypothesize that translation in specific domains like medical, law, or IT could still be challenging for the \plm as the model is less likely to have observed even sufficient monolingual datasets for these specialized domains, in contrast to the news text found in \wmt. Examples from these domains might require translating rare terminologies and carry domain-specific idiosyncrasies, which is known to pose a challenge even for a well-trained supervised neural \mt model \cite{koehn2017six}. Hence, we also evaluate \plm under these specialized out-of-domain scenarios.

\paragraph{Domain of few-shot \MakeLowercase{\incontext} examples matter.} Task-level \MakeLowercase{\incontext} examples drawn from the domain of evaluation, i.e., domain-specific, obtain on an average higher \bleu scores across the board than using examples from a distant \wmt corpus as expected (Table~\ref{tab:main_ood_eval}) in both 1-shot ($p=1$: +1.4) and 16-shot ($p=16$: +2.7) settings.  

\paragraph{Example-specific prompts significantly improve translation quality over task-level prompts.} Unlike the in-domain evaluation, retrieved and re-ranked example-specific prompts (R-\bm) improve the translation quality significantly across the board with up to 23 \bleu gain in the Law domain using just a single example as a prompt over a task-level prompt. This can be attributed to the high lexical overlap in the examples retrieved from the training data for these domains (Table~\ref{tab:bleuoverlap}). 

\paragraph{Task-level and R-\bm prompts are complementary.} Both task-level and R-\bm provide supporting information for a given test source sentence as concatenating these set of prompts improves output quality over using these methods independently, outperforming a strong \knnmt baseline on 2 out of 4 domains (Medical and IT) without requiring access to a strong base \mt model or token-level retrieval during inference. Our manual analysis suggests that the higher gain obtained in the IT domain ($+0.86$) when using both task-level and example-specific prompts can be explained by the observation that for $100$ test source sentences, there are no training examples with any lexical overlap with the test source. The task-level prompt can still elicit learned information from the \plm over using no examples for these inputs.

\section{Analysis}

\subsection{Task-level Example Selection}


\begin{figure*}[ht!]
\centering
\begin{subfigure}{0.40\textwidth}
  \includegraphics[width=\linewidth]{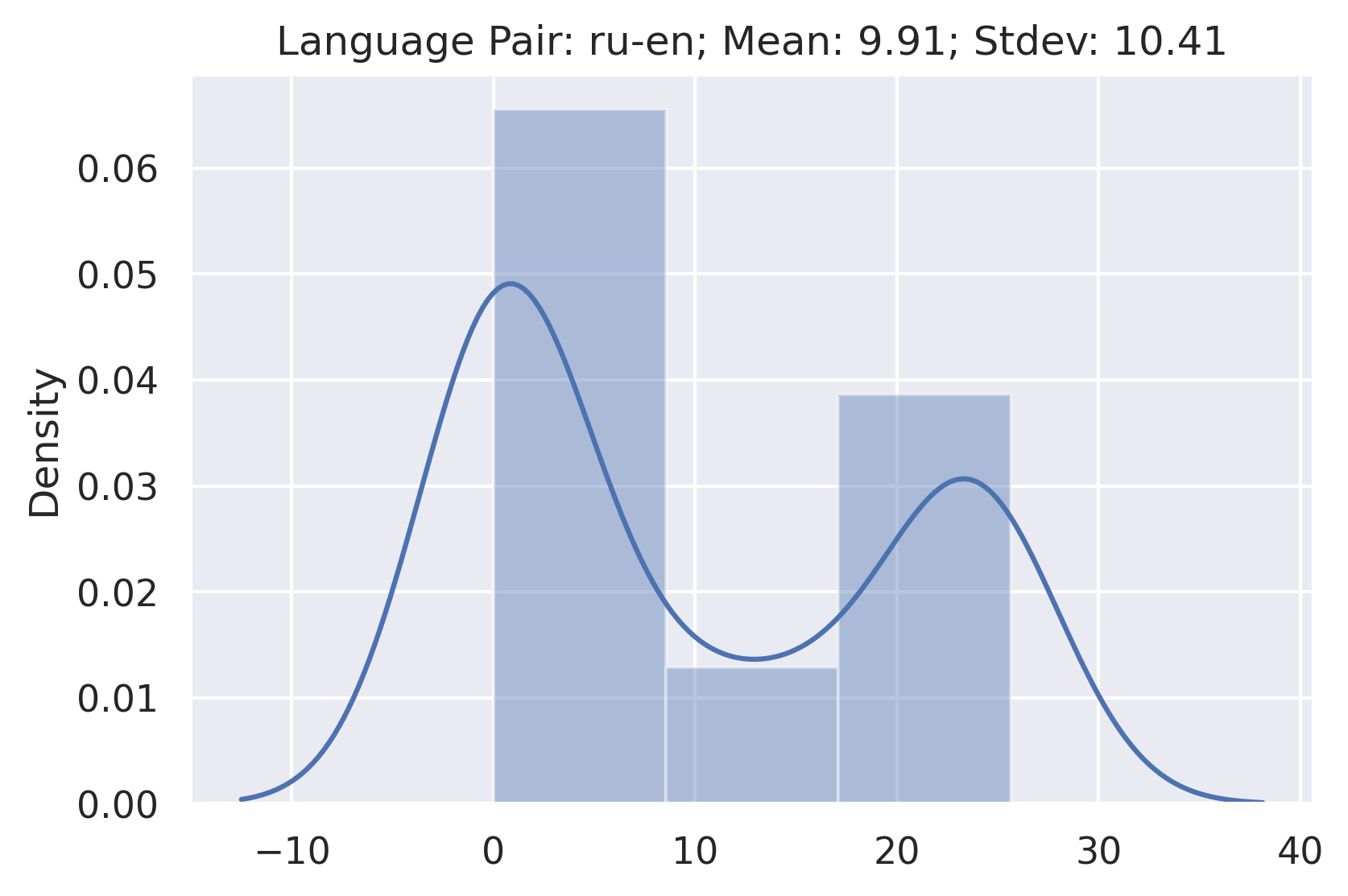}
\end{subfigure}%
\begin{subfigure}{0.40\textwidth}
  \includegraphics[width=\linewidth]{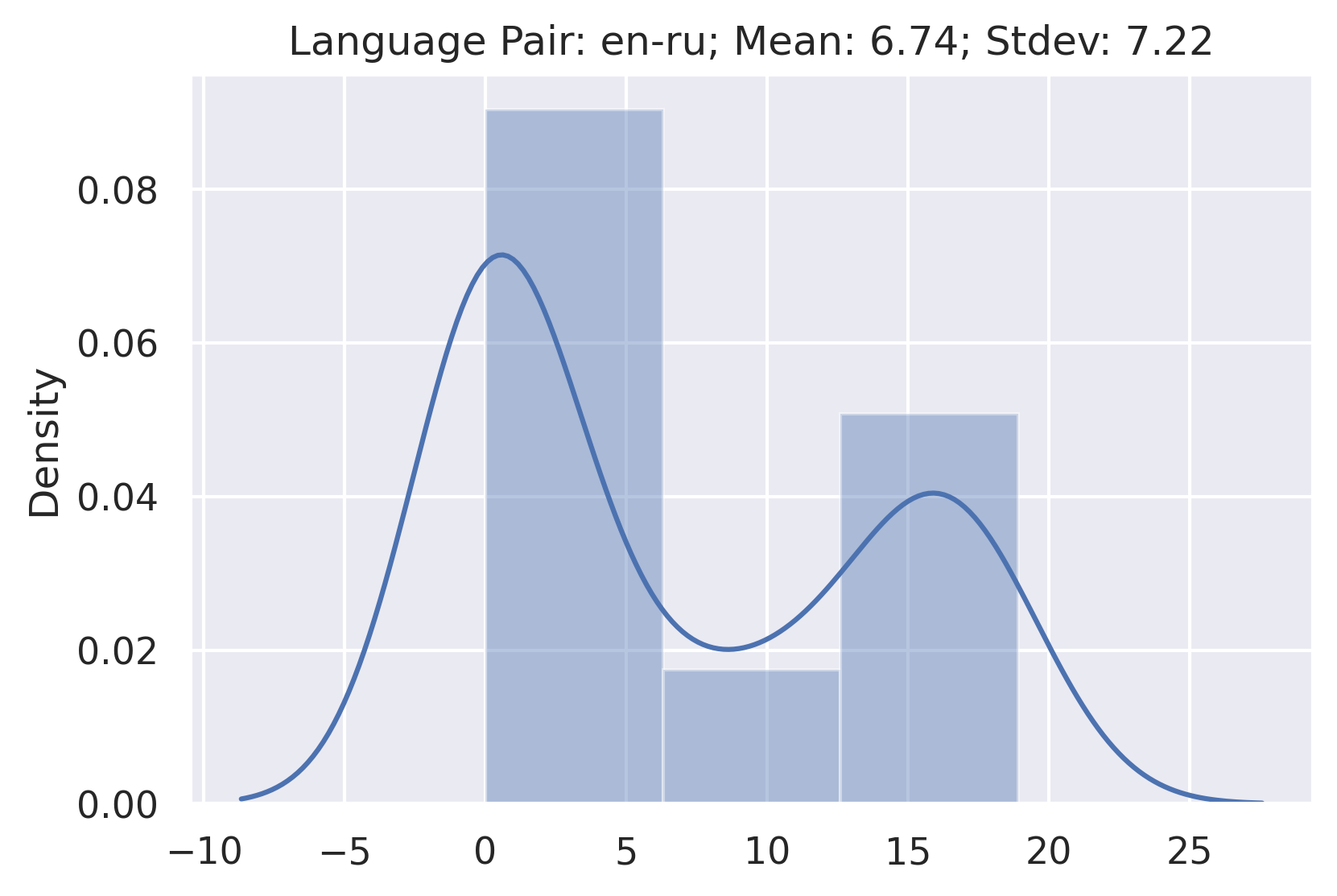}
\end{subfigure}
\begin{subfigure}{0.40\textwidth}
  \includegraphics[width=\linewidth]{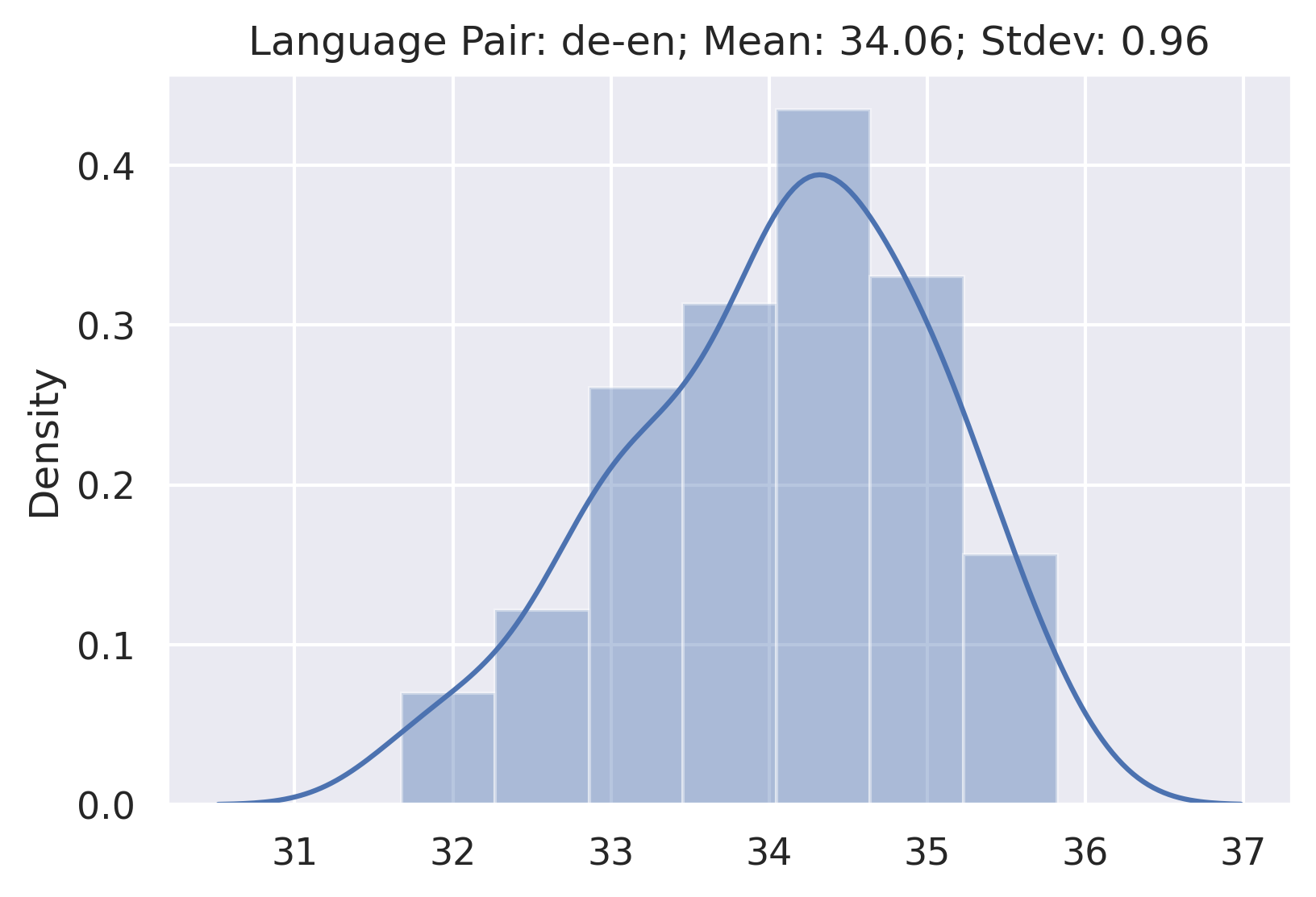}
\end{subfigure}%
\begin{subfigure}{0.40\textwidth}
  \includegraphics[width=\linewidth]{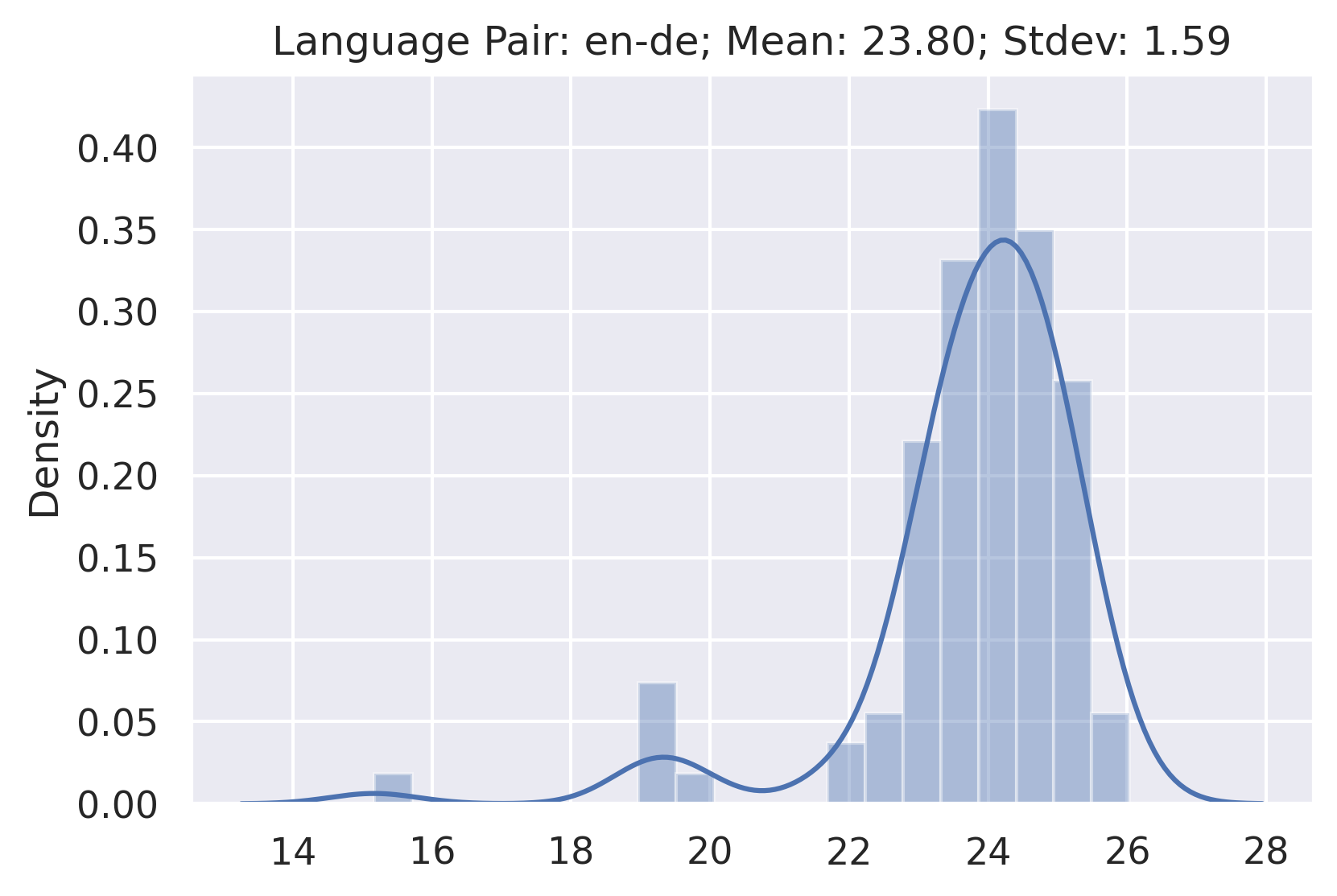}
\end{subfigure}
\caption{\bleu distribution on the WMT'18 test set for $100$ randomly sampled $1$-shot prompts from the training dataset. The same set of $100$ random $1$-shot prompts are used for x$\rightarrow$y and $y\rightarrow$x translation directions.}  \label{fig:bleudist_1shot_random}
\end{figure*}
\begin{table*}[h]
\centering
\scalebox{0.85}{
\begin{tabular}{lllll}
 \toprule
 \textbf{Features}  & \textbf{En-De} & \textbf{De-En} & \textbf{En-Ru} & \textbf{Ru-En} \\
  \midrule
     \rowcolor{gray!10}  \bf{\textit{\% (Aligned words)  }} \\
     Random (T=3) & 0.818  \small$\pm 0.009$ & 0.837 \small $\pm ~0.015$ & ~0.594 \small $\pm 0.106$ & 0.663 \small $\pm 0.051$  \\
      Task-level &  0.834 & 0.926 & 0.773 & 0.886 \\

\addlinespace[0.2cm]
     \rowcolor{gray!10}  \bf{\textit{Prism-Src}} \\
Random (T=3) &  -1.027  \small$\pm 0.068$ & -1.081  \small$\pm 0.016$ & -2.214  \small $\pm 0.139$ & -1.767  \small$\pm 0.111$ \\
 Task-level & -0.843 & -0.847 & -1.557 & -1.206\\

\bottomrule
 \end{tabular}}
\caption{Average scores obtained by top-10 1-best prompts and 10 Random 1-shot prompts on features quantifying semantic equivalence/translation quality (higher is better).}\label{tab:featurestaskprompt}
\end{table*}


\paragraph{Choice of Few-shot Examples} We show the distribution of output quality as measured by \bleu when using 100 different examples as prompts in Figure~\ref{fig:bleudist_1shot_random}. Across all four language pairs, there is a large variation in \bleu scores (up to ~20 \bleu), where noisy or unrelated prompts can lead to significantly worse output quality. Given that most existing parallel corpora are web-crawled and the quality of bitext can vary significantly across different language pairs \cite{kreutzer2022quality}, randomly sampled examples can under-estimate the translation quality attainable by prompting these \plm. 

\begin{table}[h]
\centering
\scalebox{0.85}{
\begin{tabular}{lrr}
 \toprule
 &  \multicolumn{2}{c}{ \textbf{1-shot Prompts} } \\
 & \bf{100} & \bf{1000}  \\
  \midrule
Max & 35.82 & 36.29\\
Mean & 34.06 & 29.95\\
Stdev & 0.96 & 9.55\\
\midrule
\multicolumn{3}{l}{Random 10 trials of best over 100 1-shot Prompts} \\
Mean over Max & - & 36.08 \\
Stdev over Max & - & 0.18 \\
  \bottomrule
 \end{tabular}}
\caption{Task-level example selection from 1000 1-shot Prompts on the WMT'19 development dataset. 
}\label{tab:scale1000deen}
\end{table}
\paragraph{Impact of Pool Size on Task-level Prompt Selection} We select the best task-level prompt based on the translation quality on the development set from a random sample of 100 examples (pool) as detailed in Section~\ref{subsec:task}. However, one concern regarding the selection of the best task-level prompt in this fashion could be that we might still be underestimating the \plm(s) performance, as a larger pool size could result in better output quality. We study the impact of using a larger pool size in Table~\ref{tab:scale1000deen} where increasing the number of examples from $100$ to $1000$ only leads to a gain of 0.5 points in the maximum \bleu. From the same table, we can also observe that for any subset of random $100$ few-shot examples, we can extract a task-level prompt (\bleu: ~36) with a small standard deviation in overall outpust quality ($0.18$).

\begin{figure}[h]
\centering
\begin{subfigure}{0.40\textwidth}
  \centering
  \includegraphics[width=\linewidth]{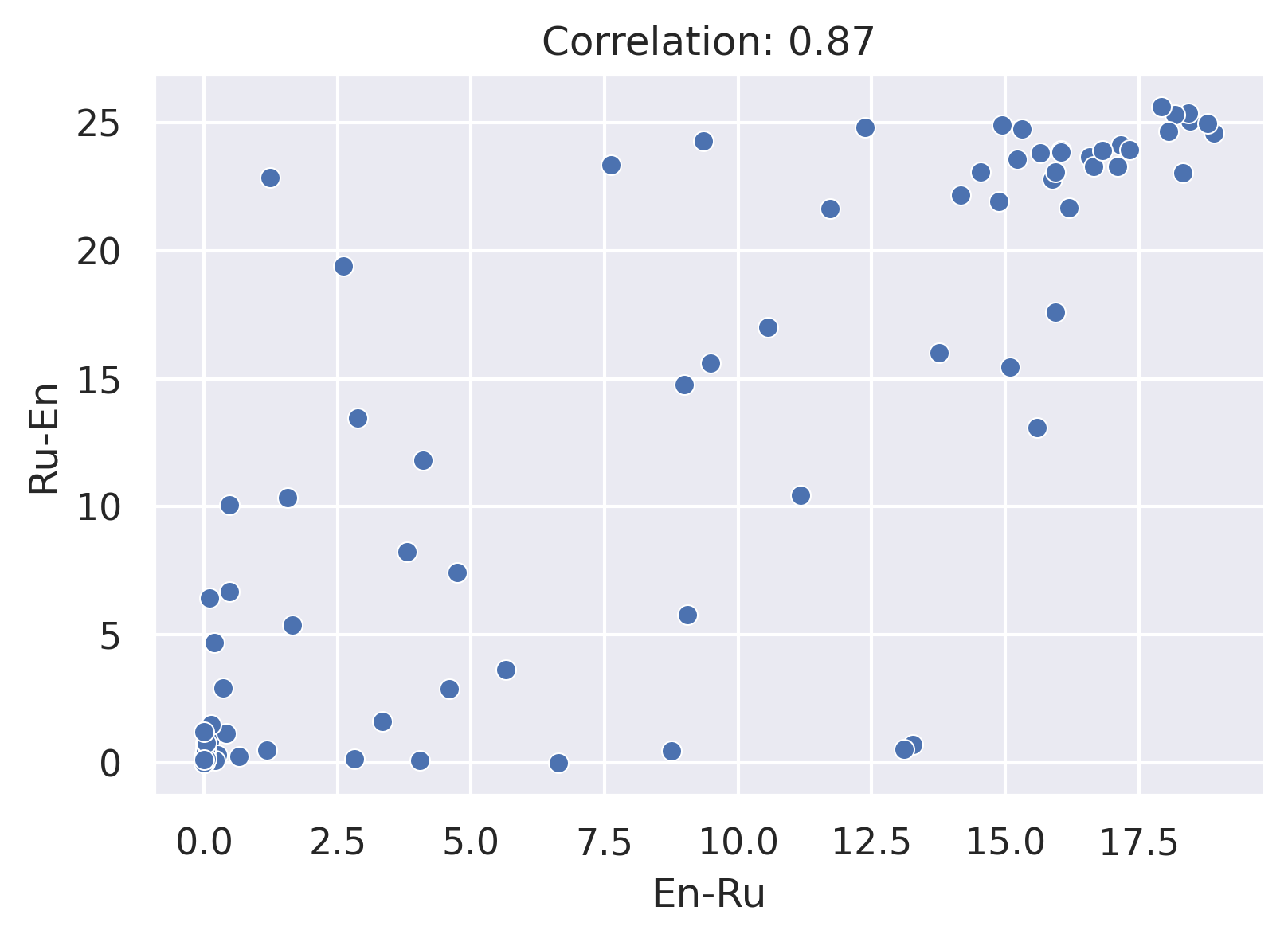}
\end{subfigure}%
\vfil
\begin{subfigure}{0.40\textwidth}
  \includegraphics[width=\linewidth]{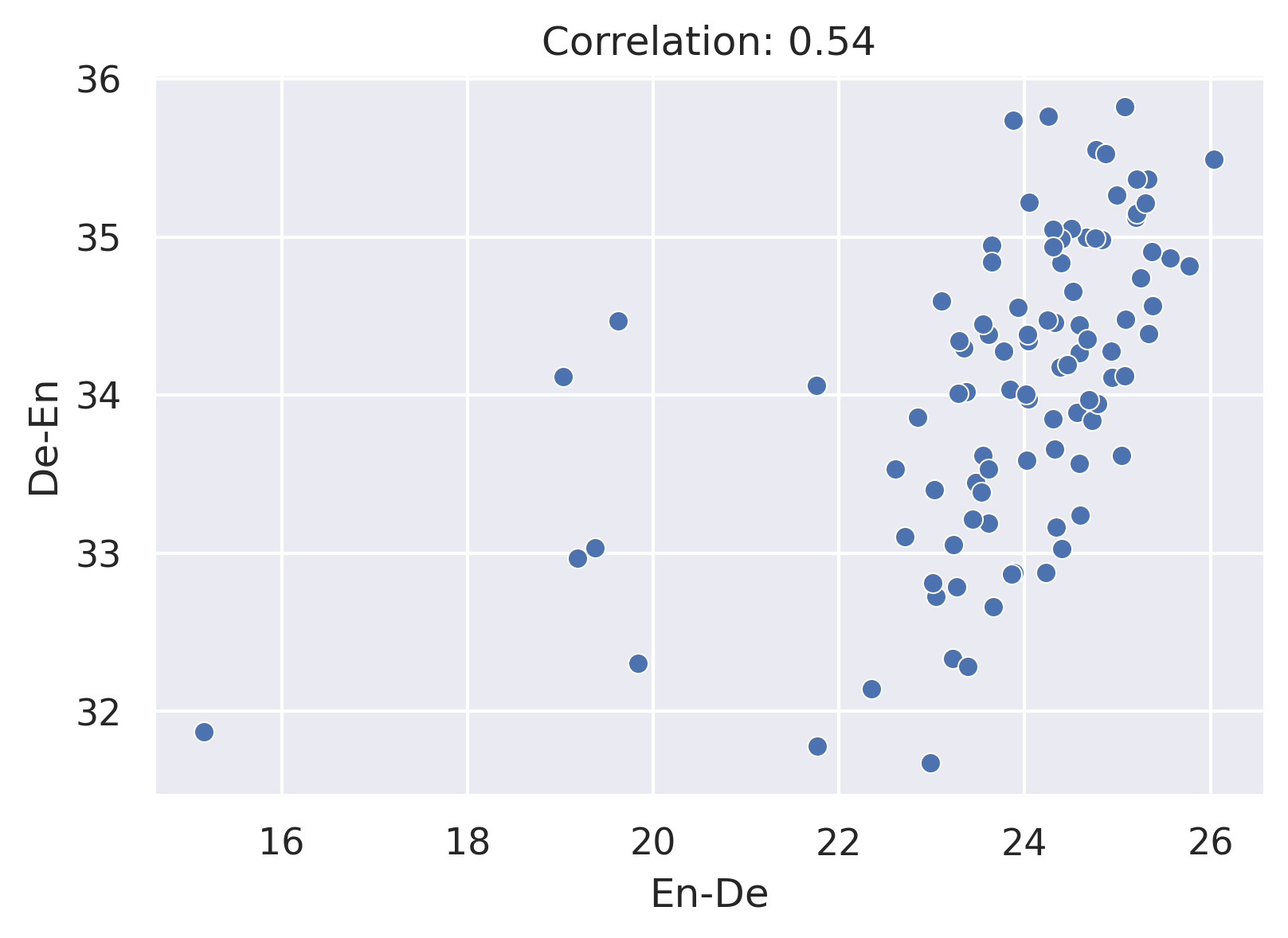}
\end{subfigure}
\caption{Best 1-shot prompts work equally well for both translation directions.}  \label{fig:bleudist_1shot_random_corr}
\end{figure}

\paragraph{Translation direction} Figure~\ref{fig:bleudist_1shot_random_corr} shows the correlation between output quality in forward ($x \rightarrow y)$ and reverse ($y \rightarrow x)$ translation directions when using 1-shot prompts --- there is a moderate to high correlation in output quality for both language pairs suggesting that the best and worst 1-shot prompts in one direction exhibit similar behavior in the opposite translation direction. 

\paragraph{Properties of good Task-level prompts} Our manual analysis on the best task-level prompts suggests that any well-formed and meaning-equivalent translation \cite{vyas-etal-2018-identifying, briakou-carpuat-2020-detecting} could make a good task-level prompt (see examples in Appendix Table~\ref{tab:taskprompts}). To quantify the meaning equivalence of the 1-best task-level prompt against random 1-shot examples, we report the percentage of aligned words between the source and reference translation (``\% Aligned words'') using fastAlign \cite{dyer-etal-2013-simple} and the probability of generating the reference translation conditioned on the source using a pre-trained multilingual \textsc{NMT} model, Prism-src \cite{thompson2020automatic, agrawal2021assessing} in Table~\ref{tab:featurestaskprompt}.\footnote{\url{https://github.com/clab/fast_align}, \url{https://github.com/thompsonb/prism}} Across all language pairs and both metrics, task-level examples achieve higher semantic similarity scores than random 1-shot examples suggesting that task-level examples are relatively more equivalent in meaning than random examples.

\begin{table}[h]
\centering
\begin{tabular}{lr}
 \toprule
 \textbf{Prompt}  & \textbf{\bleu} \\
  \midrule
Best & 32.14\\
Worst & 1.12\\
Best + Worst & 25.54 \\
Worst + Best & 31.43\\
  \bottomrule
 \end{tabular}
\caption{\bleu when using a equivalent translation against a noisy unrelated parallel example on the WMT18 De-En Development Dataset.}\label{tab:bestworse}
\end{table}
\paragraph{Impact of Noise} Table~\ref{tab:bestworse} shows the impact on translation quality when using an unrelated translation example against a meaning-equivalent or well-formed source-translation pair as a prompt: a noisy unrelated \MakeLowercase{\incontext} example leads to extremely low \bleu score of $~1$. While using both a good task-level and a noisy prompt via concatenation (best-then-worst and worst-then-best) reduces the effect of noise, the best output quality achieved when using both prompts ($31.43$) is still lower than using just the best task-level prompt ($32.14$), highlighting the importance of both prompts selection and their ordering on translation quality.

\paragraph{Impact of Ordering} To further explore the sensitivity to the order of few-shot prompts on \mt quality, we use all possible order permutations of four randomly sampled examples and the top four task-level examples as prompts ($4!$) and report the translation quality as measured by \bleu in Table~\ref{tab:promptorder}: Task-level prompts are less sensitive to prompt order, as suggested by the lower standard deviation achieved in all settings, and result in higher translation quality than randomly selected examples. Across the three different runs of randomly sampled examples, there is a significant difference in \bleu, further corroborating that the choice of \MakeLowercase{\incontext} examples and their ordering matters. 
\begin{table}[h]
\centering
 \setlength\tabcolsep{2pt}
\scalebox{0.80}{
\begin{tabular}{lllll}
 \toprule
 & \textbf{En-De} & \textbf{De-En} & \textbf{En-Ru} & \textbf{Ru-En} \\
  \midrule
 &  34.43 \small$\pm 0.25$ & 25.19 \small$\pm 0.26$& 12.48 \small$\pm 5.72$ & 15.56 \small$\pm 0.50$ \\
Random &  35.63 \small$\pm 0.48$ & 25.85 \small$\pm 0.15$& 24.99 \small$\pm 0.21$ & 19.04 \small$\pm 0.39$ \\
&  34.73 \small$\pm 0.30$ & 23.93 \small$\pm 0.28$& 10.92 \small$\pm 4.64$ & 17.91 \small$\pm 0.07$ \\
\midrule
 Optimized & 35.95 \small$\pm 0.24$  & 26.98 \small$\pm 0.15$  & 25.85 \small$\pm 0.11$  & 19.96 \small$\pm 0.24$ \\
\bottomrule
 \end{tabular}}
\caption{\bleu over all 24 permutations of 3 seeds of 4 randomly selected and top 4 task-level prompts. }\label{tab:promptorder}
\end{table}


\begin{table*}[h]
\centering
\scalebox{0.80}{
\begin{tabular}{lcccc}
 \toprule
 Dataset & Avg. \bleu($I_x$, x) & Corr(\bleu($\hat{y}$, y),\bleu($I_x$, x)) & Avg. \bleu($I_y$, y) & Corr(\bleu($\hat{y}$, y),  \bleu($I_y$, y))\\
  \midrule
Medical & 35.785 & 0.593 & 32.101 & 0.777\\
Law & 34.982 & 0.677 & 34.349 & 0.786\\
IT & 25.196 & 0.497 & 19.382 & 0.669 \\
Koran & 36.033 & -0.016 & 10.364 & 0.676\\
\bottomrule
 \end{tabular}}
\caption{Correlation between the degree of overlap as measured by \bleu and the translation quality of the outputs, $\bleu (\hat{y}, y)$, across different domains when using the top-1 prompt retrieved using \bm. $I_x$ and $I_y$ are the sources and the reference translations in the \bm examples respectively.}\label{tab:bleuoverlap}
\end{table*}

\subsection{Informativeness of Example-specific Prompts }
To understand the benefit of retrieved examples in the out-of-domain evaluation, we measure the lexical overlap between the test input ($x$, $y$) and the prompts ($I_x, I_y$) using \bleu (Avg. \bleu($I_x$, x), Avg. \bleu($I_y$, y)), where $I_x$ and $I_y$ are the sources and target translations of the retrieved in-context examples. We also  report the correlation against the translation quality $\bleu (\hat{y}, y)$. Table~\ref{tab:bleuoverlap} shows that the source lexical overlap is a good indicator of the informativeness of a prompt for 3 out of 4 domains, with Koran as an exception. For Koran, while the retrieved sentences have a high overlap with the source (36.03), the target associated with the prompts ($I_y$) does not get high \bleu with the reference (10.36) compared to other domains. We hypothesize that this might be due to a bias in the reference translations towards a particular output style. We provide an analysis of the impact of this phenomenon on \mt quality in Section~\ref{sec:outputanalysis}.

\section{Output Analysis} \label{sec:outputanalysis}

We report two interesting findings when prompting \plm with task-level and example-specific prompts:

\paragraph{Stylistic Outputs} One advantage of using a single task-level \MakeLowercase{\incontext} example to prompt the \plm is that it allows us to systematically study how the choice of prompt influences the style of the generated translation. Table~\ref{tab:styleexample} illustrates one such example: we can observe that as the prompt includes a \textit{contraction} (``we are'' vs. ``we're''), the outputs generated by the \plm also include contractions and can be incorrectly penalized by \bleu while being meaning equivalent.

\begin{table}[h]
    \centering
        \renewcommand\tabularxcolumn[1]{m{#1}}
        \renewcommand\arraystretch{1.1}
    \begin{tabularx}{\columnwidth}{*{1}{>{\arraybackslash}X}}
    \textbf{Prompt:} Wegen des heißen Sommers fangen wir erst spät an. = Because of the hot summer, we're late getting started. \\ 
  \textbf{Source:} Ja, ich bin sehr zufrieden mit dem Auftritt. \\ 
  \textbf{Reference:} Yes, I am very happy with the performance. \\
   \textbf{\plm Output:} Yes, I'm very satisfied with the performance. \\
   \addlinespace[0.1cm]
   \textbf{Source:} Es ist eine andere Unternehmenskultur. \\ 
   \textbf{Reference:} It is a different corporate culture. \\
   \textbf{\plm Output:} It's a different corporate culture. \\
    \end{tabularx}  
    \caption{Outputs mimic the style of the prompt.} \label{tab:styleexample}
\end{table}

\paragraph{Template-based \mt} 

Template-based translation in medical, legal, it, or e-commerce domain can be preferable as they reduce the risk of generating errors in automatically generated translations. We present some examples in Table~\ref{tab:templatetranslation} on how \plm can seamlessly use retrieved prompts to synthesize a translation from the template provided.  

\begin{table}[h]
    \centering
        \renewcommand\tabularxcolumn[1]{m{#1}}
        \renewcommand\arraystretch{1.1}
    \begin{tabularx}{\columnwidth}{*{1}{>{\arraybackslash}X}}
    \textbf{Prompt:} WIE IST SINGULAIR ANZUWENDEN? = HOW TO TAKE SINGULAIR  \\ 
  \textbf{Source:} WIE IST EVOLTRA ANZUWENDEN? \\ 
   \textbf{\plm Output:}  HOW TO TAKE EVOLTRA \\
   \addlinespace[0.1cm]
    \textbf{Prompt:} Zeigt die aktuelle Datei mit Opera an. = View the current file with Opera. \\ 
   \textbf{Source:} Zeigt die aktuelle Datei mit Lynx an (Textbasierter Browser). \\ 
   \textbf{\plm Output:} View the current file with Lynx (Text-based browser). \\
    \end{tabularx}  
    \caption{Outputs follow the template of the prompt.} \label{tab:templatetranslation}
\end{table}

\subsection{Size of the Datastore}

\begin{figure}[h]
\centering
\includegraphics[width=0.9\linewidth]{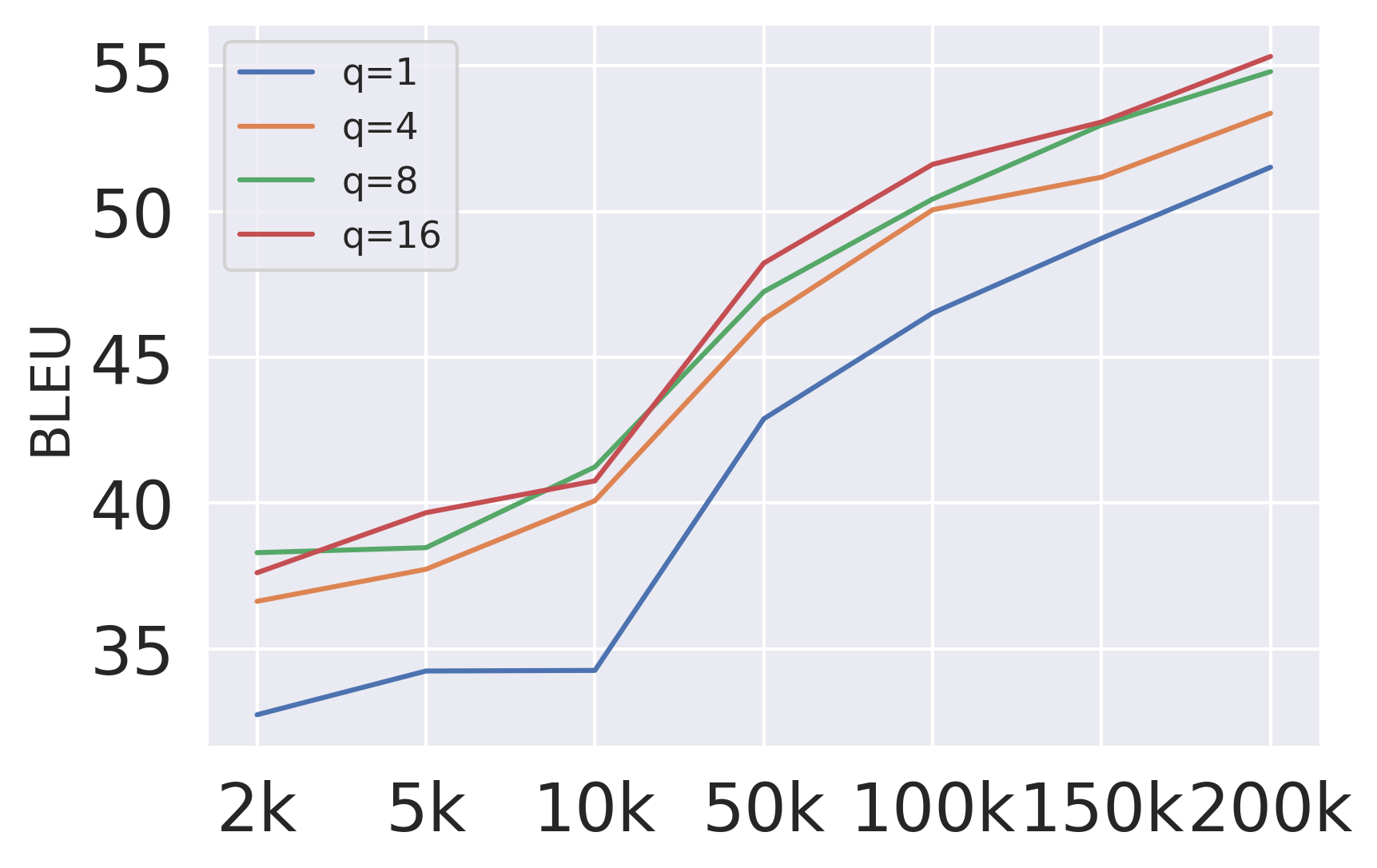}
\caption{\bleu on the Medical dataset when varying the size of the datastore and the number of \bm \MakeLowercase{\incontext} examples.} \label{fig:dstoresize}
\end{figure}

Figure~\ref{fig:dstoresize} shows \bleu when varying the size of the datastore used to retrieve similar \MakeLowercase{\incontext} examples using \bm on the Medical dataset. As the size of the datastore increases, the likelihood of retrieving a more similar example increases. However, similar output quality in \bleu can be achieved by using multiple \MakeLowercase{\incontext} examples when a smaller in-domain datastore is available as multiple examples can provide better coverage of the source terms --- \bleu@q=16 with a datastore size of 100k is equivalent to \bleu@q=1 with twice as many examples (200k). 

%

\section{Related Work}

\paragraph{\incontext Learning for \mt}
\citet{garcia2022using} use natural language prompts (e.g. Translate to \{language\_name\}: \{text\}) to control the target language in multilingual \mt and investigate the impact of scale, number of languages, and their similarity for this phenomena. \citet{wang2022training} utilize \bm retrieved training examples in a supervised fashion to learn from similar examples during training. Contrary to prior work, we utilize \textit{similar} examples to form a textual prompt which is used to guide the generation of translation output during inference and systematically study the properties of good in-context examples for \mt. 

\paragraph{Domain Adaptation for \mt} Prior work on domain adaptation for machine translation uses out-of-domain bilingual or monolingual datasets to improve the translation quality of a pre-trained neural sequence-to-sequence \mt model either during training \cite{luong-manning-2015-stanford, freitag2016fast, wang-etal-2017-instance} or inference \cite{zheng2021non, khandelwal2020nearest}. Similar to past work, our work utilizes out-of-domain datasets during inference to adapt a pre-trained generative language model to improve the translation quality on unseen domains. However, our approach does not rely on creating a domain-specific token-level datastore, but directly uses \text{similar} examples to provide additional context, hence is more compute and memory efficient.

\paragraph{Prompt Selection}
The importance of selecting good \MakeLowercase{\incontext} examples and their impact on downstream \nlp task performance has been studied in prior work \cite{liu-etal-2022-makes, lu2022fantastically, jiang-etal-2020-know, min2022rethinking, zemlyanskiy2022generate, rubin2021learning, liu2022few}. However, how these examples and their properties impact \mt quality remains unexplored, which we investigate in our work.

\section{Conclusion}
We investigate the choice of \MakeLowercase{\incontext} examples selection for \mt in both in-domain and out-of-domain settings. We propose a novel recall-based re-ranking approach to utilize similar training examples as prompts and show their efficacy across multiple datasets and domains. Our findings show that task-level prompts can provide a complementary advantage to example-specific prompts, outperforming a strong \knnmt baseline in 2 out of 4 out-of-domain datasets while being memory and compute efficient. Our manual analysis of the generated outputs reveals that the \plm can mimic the style of the \MakeLowercase{\incontext} examples provided and can be used for template-based translation synthesis. These results open space for future research to evaluate the potential of generating diverse and style-specific outputs for \mt.

\bibliography{anthology,custom}

\appendix
\clearpage

\section{Statistics of Datasets}
Table~\ref{tab:ablation_datastats} includes statistics of training, development and test sets used for the experiments discussed in the paper.
\begin{table}[h]
\centering
\begin{tabular}{lrrr}
 \toprule
 \textbf{Dataset}  & \textbf{Train}  & \textbf{Dev} & \textbf{Test} \\
  \midrule
WMT-19 (de) & $42$M & 2998 & 2000\\
WMT-19 (ru) & $10$M & 3000 & 2000\\
  \addlinespace[0.2cm]
  \bf{\textit{Multi-Domain}} \\
  Medical & $248$K & 2000& 2000\\
  Law & $467$K & 2000& 2000\\
  IT & $223$K & 2000& 2000\\
  Koran & $17$K & 2000& 2000\\
  \bottomrule
 \end{tabular}
\caption{Dataset Statistics.}\label{tab:ablation_datastats}
\end{table}

\section{Compute Infrastructure \& Run time}

Each experiment is run on a single Nvidia Tesla V100 Volta GPU machine with 32G Ram. A single inference experiment on $2000$ test examples using \xglm with $16$ in-context examples takes around 3-4 hrs to complete.

\section{Results using Second Metric: Comet}

We report translation quality using Comet \cite{rei-etal-2020-comet} in Tables~\ref{tab:comet_ind_eval} and \ref{tab:comet_ood_eval}. We use the \texttt{eamt22-cometinho-da} model \cite{rei-etal-2022-searching} to generate the scores as it was shown to achieve higher correlations with human judgments than lexical overlap metrics while being computationally efficient. Our re-ranking strategy (with $q_{max}=16$) consistently performs the best across the board except for Koran, outperforming strong \knnmt baselines on the multi-domain test set in 3 out of 4 settings. Adding a task-level prompt to 16 R-\bm prompts via concatenation further improves quality in 5 out of 8 settings.  

\section{Hyperparameter Search}
\subsection{Order of \bm Retrieved Examples }

We report the \bleu when using two different orderings of example-specific prompts on the development set for the medical domain. Ordering the examples with the most similar examples on the left attains higher \bleu than the right-to-left order.  We note that the trend could vary depending on the noise in the training dataset, the degree of similarity, and the number of retrieved examples. We leave the exploration of the ordering of example-specific prompts to future work.
\begin{table}[h]
\centering
\begin{tabular}{lr}
 \toprule
 \textbf{$\lambda$}  & \textbf{\bleu} \\
  \midrule
Left-to-right & 56.84 \\
Right-to-left & 54.97 \\
  \bottomrule
 \end{tabular}
\caption{\bleu using twi different orderings of the top-16 example-specific \bm prompts on the Medical development Set.}\label{tab:order}
\end{table}

\subsection{Choice of $\lambda$, Threshold}

Table~\ref{tab:hyp_lambda} shows the \bleu and the average number of \MakeLowercase{\incontext} examples selected when varying $\lambda$ and the threshold described in Section~\ref{subsec:example}. We select $\lambda=0.1$ and threshold value of 1.0 as it achieves the best \bleu on the Medical development set as shown below:
\begin{table}[h]
\centering
\begin{tabular}{lrrr}
 \toprule
 \textbf{$\lambda$} & \textbf{Threshold}  & \textbf{\bleu} & \textbf{Avg. \# of Examples} \\
  \midrule
0.1 & 0.1 & 54.55 & 14.16 \\
& 1.0 & \bf{54.56} & 12.73 \\
 & 5.0 & 53.35 & 8.83\\
0.3  & 0.1 & 54.47 & 15.06 \\
& 1.0 & 54.51 & 14.28 \\
 & 5.0 & 53.98 & 10.32 \\
0.5  & 0.1 & 54.44 & 15.44 \\
& 1.0 & 54.39 & 15.10 \\
 & 5.0 & 54.44 & 11.85 \\
  \bottomrule
 \end{tabular}
\caption{\bleu using different values of $\lambda$ and threshold on the Medical Development Set ($q_{max}=16)$.}\label{tab:hyp_lambda}
\end{table}

\section{Example Task-Level Prompts}
Table~\ref{tab:taskprompts} shows the best task-level in-context example selected by our method described in \S~\ref{subsec:task} and the respective \bleu scores on the development set for the German-English and Russian-English tasks.

\begin{table*}[h]
    \centering
        \renewcommand\tabularxcolumn[1]{m{#1}}
        \renewcommand\arraystretch{1.3}
    \begin{tabularx}{\textwidth}{*{1}{>{\arraybackslash}X}}
  \textbf{German:} Beispielsweise der Änderungsantrag zu Artikel 5 in der Stellungnahme des Ausschusses für Landwirtschaft und ländliche Entwicklung weist klar und deutlich darauf hin, dass die Verschlechterung der Qualität des Bodens lokale oder regionale Ursachen und Wirkungen hat und daher unbedingt nationale statt europäischer Maßnahmen ergriffen werden müssen. \\
  \textbf{English:}  For example, the amendment to Article 5 in the opinion of the Committee on Agriculture and Rural Development clearly indicates that the degradation of the soil has local or regional causes and effects and it is therefore essential to adopt national as opposed to European measures. \\
    \textbf{Development BLEU: } 35.82 \\
     \addlinespace[0.2cm]
\textbf{Russian:} \foreignlanguage{russian}{Если ваш браузер возвращает ранее сохраненный ``cookie'', то управляющий им поставщик имеет возможность соединить актуальное посещение пользователя с предыдущими посещениями, но только в отношении своего содержания. } \\
 \textbf{English:}  If the browser sends back an earlier saved cookie, then the service managing these can connect to the user\'s earlier visit, but only in respect of their own content. \\
    \textbf{Development BLEU: } 25.63 \\
    \end{tabularx}  
    \caption{Best task-level prompt For De-En and Ru-En Language Pairs according to the \bleu score on the development set. } \label{tab:taskprompts}
\end{table*}


\begin{table*}[h]
\centering
\scalebox{0.90}{
\begin{tabular}{lcrrrr}
 \toprule
\bf{ Method}& $\bf{p+q_{max}}$ &  {\bf{En-De}}  & {\bf{De-En}}& {\bf{Ru-En}}   &  {\bf{En-Ru}}  \\ 
  \midrule
Task-level & \colorbox{pink}{$1+0$}  & 0.354 &  0.403 & 0.428 & 0.626\\
 \addlinespace[0.1cm]
 \bm& \colorbox{pink}{$0+1$} & 0.107 & 0.149 & 0.139 & 0.346 \\
 \addlinespace[0.1cm]
R-\bm & \colorbox{pink}{$0+1$}     & 0.204  & 0.249  & 0.244 & 0.413 \\

\addlinespace[0.1cm]
\midrule
\addlinespace[0.1cm]

Random-Avg & \colorbox{paleyellow}{$16+0$}  & 0.387  & 0.391 & 0.424 & 0.636 \\
      \addlinespace[0.1cm]
Task-level  & \colorbox{paleyellow}{$16+0$}  & 0.389 &  0.381 & 0.440 & 0.662 \\
        \addlinespace[0.1cm]
 \bm   & \colorbox{paleyellow}{$0+16$} & 0.423 &  0.410 & 0.434 & 0.673 \\
    \addlinespace[0.1cm]
  R-\bm  & \colorbox{paleyellow}{$0+16$} & 0.438 & 0.420 & 0.444 & 0.677 \\
  
\addlinespace[0.1cm]
\midrule
\addlinespace[0.1cm]

   R-\bm  & \colorbox{mint}{$0+17$} & \bf{0.440} &  0.421  & \bf{0.448} & 0.676 \\
   \addlinespace[0.1cm]
Task-level + R-\bm &   \colorbox{mint}{$1+16$}  & 0.434 &  \bf{0.430} & 0.447 & \bf{0.694} \\
 
  \bottomrule
 \end{tabular}
 }
\caption{Comet Scores on \textsc{WMT}'19 test sets.}\label{tab:comet_ind_eval} 
\end{table*}
\begin{table*}[h]
\centering
\scalebox{0.85}{
\begin{tabular}{llcrrrr}
 \toprule
\bf{Method}& \bf{Corpus} & $\bf{p+q_{max}}$  & {\bf{MEDICAL}} & {\bf{LAW}} & {\bf{IT}}   &  {\bf{KORAN}}  \\ 
\midrule
  \multicolumn{2}{l}{Results from \citet{jiang2022towards}} \\
  Vanilla \knnmt & - & -  & 0.548 & 0.662 & 0.531 & -0.014 \\
  Their model & - & - & 0.578 & 0.703  &  0.585 & 0.047 \\
  \midrule
\multirow{2}{*}{Task-level} &  Domain-specific & \multirow{2}{*}{\colorbox{pink}{$1+0$}} & 0.314 & 0.320& 0.240 & -0.068\\
& \textsc{WMT}  &  & 0.277 & 0.345 & 0.146 & -0.113 \\
R-\bm &  Domain-specific & \colorbox{pink}{$0+1$} &  0.464 & 0.553 & 0.389 & -0.216 \\

\addlinespace[0.1cm]
\midrule
\addlinespace[0.1cm]

\multirow{2}{*}{Task-level} & Domain-specific  & \multirow{2}{*}{\colorbox{paleyellow}{$16+0$}} & 0.369 & 0.365 & 0.222 & \bf{-0.047} \\
&  \textsc{WMT} &  & 0.297 & 0.399  & 0.098 & -0.131\\
   \addlinespace[0.1cm]
 R-\bm& Domain-specific  & \colorbox{paleyellow}{$0+16$} &  0.697 & 0.697 & 0.666 & -0.105 \\

\addlinespace[0.1cm]
\midrule
\addlinespace[0.1cm]
     R-\bm   & \multirow{2}{*}{Domain-specific} & \colorbox{mint}{$0+17$} & 0.699 & 0.697 & 0.667 & -0.104\\
     \addlinespace[0.1cm]
 Task-level + R-\bm &      &    \colorbox{mint}{$1+16$} & \bf{0.701} & \bf{0.699} & \bf{0.721} &  -0.095 \\ 
   \addlinespace[0.1cm]

  \bottomrule
 \end{tabular}
 }
\caption{Comet Scores on the Multi-Domain Test Set. 
}\label{tab:comet_ood_eval} 
\end{table*}



\end{document}